\documentclass{article}[twocolumn]
\usepackage[T1]{fontenc}
\usepackage[utf8]{inputenc}
\usepackage{graphicx}
\usepackage{amsmath}
\usepackage{multicol}
\usepackage{csquotes}
\usepackage{booktabs}
\usepackage{xcolor}
\usepackage{hyperref}
\usepackage{subcaption}
\usepackage{amsfonts}
\usepackage{amsthm}
\usepackage{stmaryrd}
\usepackage{latexsym}
\usepackage{tikz}
\usepackage[inline]{enumitem}
\usepackage{natbib}
\usepackage{arxiv}

\setcitestyle{authoryear,round,citesep={;},aysep={,},yysep={;}}

\theoremstyle{definition}

\title{Robust Explanations Through Uncertainty Decomposition: A Path to Trustworthier AI}

\author{
 Chenrui Zhu\\
 CNRS, Universit\'e de technologie de Compi\`egne\\
 UMR CNRS 7253 Heudiasyc, France \\
 \texttt{chenrui.zhu@utc.fr} \\
 \And
 Louenas Bounia\\
 CNRS, Université Sorbonne Paris Nord \\
 UMR CNRS 7030 LIPN, France \\
 \texttt{bounia@lipn.univ-paris13.fr} \\
 \And
 Vu‑Linh Nguyen\\
 CNRS, Universit\'e de technologie de Compi\`egne\\
 UMR CNRS 7253 Heudiasyc, France \\
 \texttt{vu-linh.nguyen@utc.fr} \\
 \And
 Sébastien Destercke\\
 CNRS, Universit\'e de technologie de Compi\`egne\\
 UMR CNRS 7253 Heudiasyc, France \\
 \texttt{sebastien.destercke@utc.fr} \\
 \And
 Arthur Hoarau\\
 CNRS, Universit\'e de technologie de Compi\`egne\\
 UMR CNRS 7253 Heudiasyc, France \\
 \texttt{arthur.hoarau@utc.fr} \\
}

\begin{document}

\maketitle


\begin{abstract}
Recent advancements in machine learning have emphasized the need for transparency in model predictions, particularly as interpretability diminishes when using increasingly complex architectures. In this paper, we propose leveraging prediction uncertainty as a complementary approach to classical explainability methods. Specifically, we distinguish between aleatoric (data-related) and epistemic (model-related) uncertainty to guide the selection of appropriate explanations. Epistemic uncertainty serves as a rejection criterion for unreliable explanations and, in itself, provides insight into insufficient training (a new form of explanation). Aleatoric uncertainty informs the choice between feature-importance explanations and counterfactual explanations. This leverages a framework of explainability methods driven by uncertainty quantification and disentanglement. Our experiments demonstrate the impact of this uncertainty-aware approach on the robustness and attainability of explanations in both traditional machine learning and deep learning scenarios.
\end{abstract}


\section{Introduction}

Explainable Artificial Intelligence (XAI) \citep{Miller19,das2020opportunities,Anchor18,Lime16,Lundberg17,Dhurandhar2018,JoaoLogicXAI, IgnatievNM19,DarwicheH20,Audemardetal22} and Uncertainty Quantification (UQ) \citep{huang2023,hullermeier22,kendall2017,Nguyen2022,senge2014} are two important research directions seeking insights about predictive models. While XAI typically seeks insights about how the predictive models make predictions and why particular predictions are made, UQ typically focuses on quantifying how (un)certain the model's predictions are. Clearly, both the explanations given by XAI methods and degrees of uncertainty can be seen as ways to make different aspects of the predictive models more transparent. Degrees of uncertainty are built on top of rigorous uncertainty theories to be context-dependent, yet interpreting them to the end-user might be difficult. In contrast, XAI mechanisms are typically more human-friendly, but studies usually focus on a single, context-independent mechanism, by which we mean that the same type of explanations is outputted whatever the considered instance. To our knowledge, there is a lack of approaches seeking to adapt the explanations to the context, using different notions of explanations to be more informative and relevant. Mixing the complementary nature of uncertainty quantification and notions of explanations seems a natural way 
to construct guidelines to provide context-dependent, human-friendly explanations, that can then help to enhance their acceptability among users.

In the UQ literature, many attemps \citep{huang2023,hullermeier22,kendall2017,senge2014} have been devoted to distinguishing and quantifying two types of uncertainty~\citep{Hora1996}: \emph{aleatoric uncertainty} (AU) is rooted in the stochastic nature of the data-generating process and measure a degree of ambiguity, and \emph{epistemic uncertainty} (EU) is associated with a lack of knowledge and measure whether the model is informed or not. These degrees of uncertainty have been leveraged for different purposes and tasks: when seeing regions with high degrees of epistemic uncertainty, additional annotated data from these regions are sought to enhance the representativeness of the current observed data \citep{Nguyen2022}. However, when seeing regions with high degrees of aleatoric uncertainty, it may be more beneficial to seek either more flexible hypothesis spaces or additional features to better separate classes \citep{hoarau2025}.  



In the classification setting considered in this paper, and generally in supervised learning, XAI approaches can be categorized into local and global methods. While global methods typically quantify the model's global behavior \citep{Lundberg2019ExplainableAF}, local  methods aim to explain the local prediction of an instance $x$ classified by a predictive model $h$. We will focus on local XAI methods, which may be more readily combined with the notion of uncertainty degrees, often also defined locally. These methods are generally divided into two main categories: (\emph{i}) model-agnostic approaches, such as LIME~\citep{Lime16}, SHAP~\citep{Lundberg17}, or counterfactual explanations \citep{Dhurandhar2018}, and (\emph{ii}) formal approaches using model-specific features, such as various prime implicant concepts (\emph{cf.}, \citep{JoaoLogicXAI, IgnatievNM19, DarwicheH20, Audemardetal22}) or sufficient reasons \citep{ShihCD18}. However, those mechanisms are often considered separately, and to our knowledge there is a lack of methods that seek to combine those explanations and adapt them to contextual information such as the instance values. 


The goal of this paper is to explore how uncertainty quantification and XAI approaches can be combined to produce more relevant explanations, producing ample empirical evidence that there is benefit in epxloring their complementarity.

\subsection*{Related work}\label{section:related}
The intersection of explainability and uncertainty quantification in machine learning has recently gained momentum. The authors of ~\citep{Slack2020} proposed uncertain explanations in the form of distributions, allowing for the construction of confidence intervals and the use of uncertain explanations. The work of ~\citep{pmlr-marx23a} focuses on providing a set of possible explanations rather than a single one, leveraging imprecision in the interpretation process. In~\citep{LOFSTROM2024123154}, the authors explore model calibration and construct calibrated factual and counterfactual explanations, while ~\citep{wickstrom2024} focuses on pixel importance based on model uncertainty in image explanation tasks. However, all of these approaches aim at estimating the uncertainty of explanations, thus aiming at augmenting existing explanation mechanisms rather than using the synergies of UQ and XAI. 

Unlike these recent methods, we focus on model uncertainty as an explanation in itself, combined with other mechanisms. The recent work of~\citep{THUY2024330} is  the most closely related to our article. In their study, the authors propose to decompose model uncertainty and use it as an explanation. However, their study is quite limited, as (i) their analysis only concerns neural networks and two families of models for uncertainty quantification: Bayesian Neural Networks and Deep Ensembles\footnote{It has been shown that one can be used to approximate the other~\citep{Mobiny2021}.}, and they also set aside XAI approaches and do not explore UQ/XAI interactions.
In our work, we demonstrate that model uncertainty can serve as an explanation in itself but, more importantly, can complement classical explainability methods, particularly feature-importance explanations and counterfactual explanations. 

\begin{figure}
    \centering
    \raisebox{0.35cm}{\begin{subfigure}[t]{0.24\linewidth}
    \includegraphics[width=\linewidth]{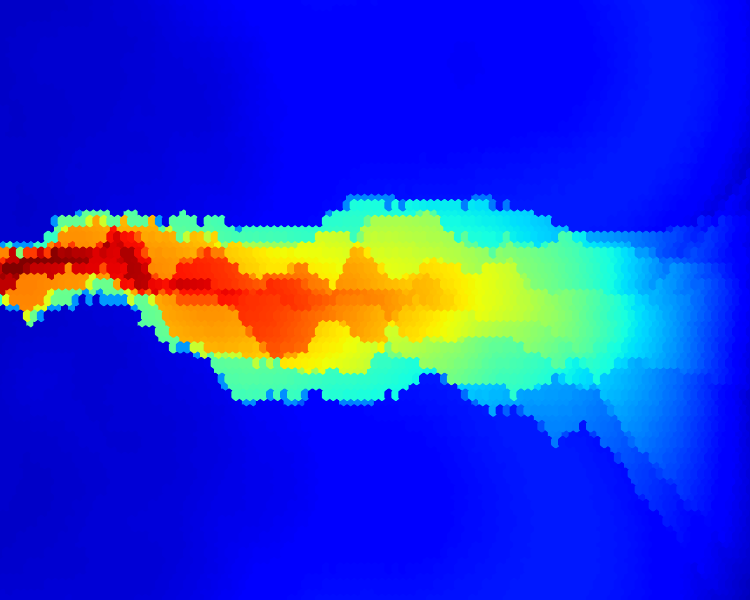}
    \end{subfigure}}
    \hspace{0.2cm}
    \begin{subfigure}[t]{0.26\linewidth}
    \includegraphics[width=\linewidth]{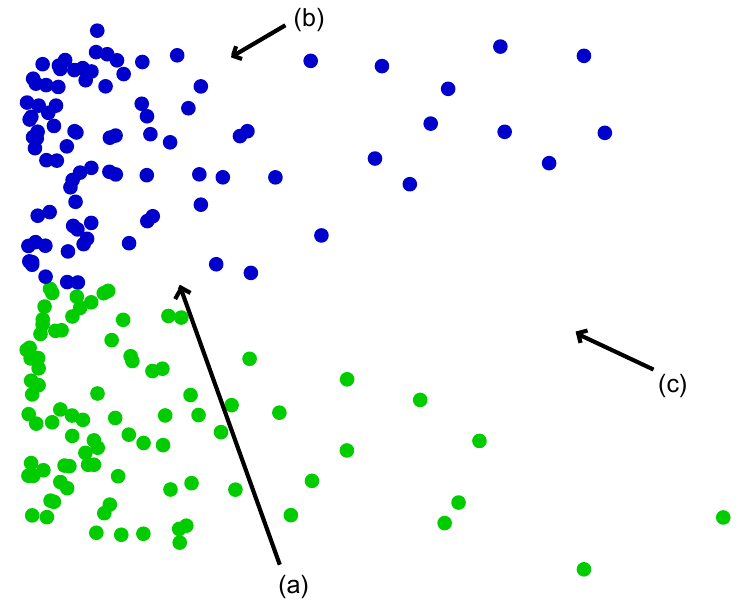}
    \end{subfigure}
    \hspace{0.2cm}
    \raisebox{0.35cm}{\begin{subfigure}[t]{0.24\linewidth}
    \includegraphics[width=\linewidth]{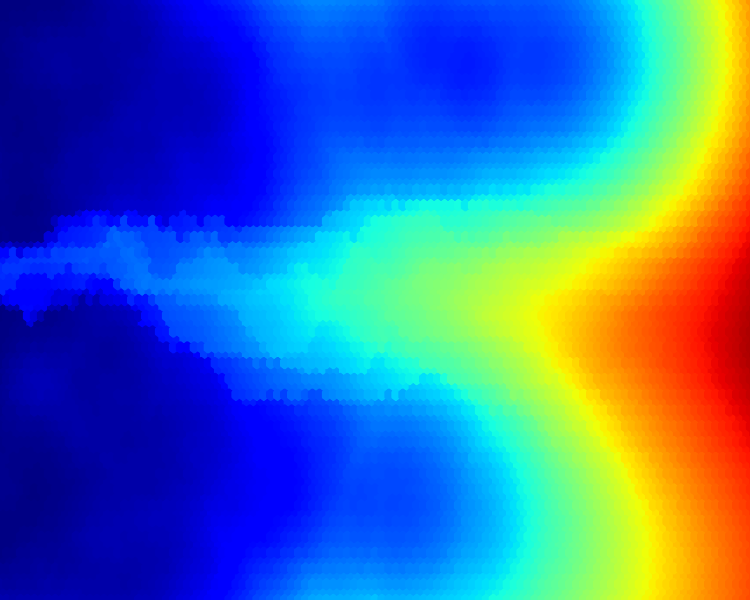}
    \end{subfigure}} 
    \caption{Uncertainty decomposition, with aleatoric uncertainty on the left (inherent difficulty of the classification task) and epistemic uncertainty on the right (lack of training data) for a toy dataset. Three areas are distinguished: (a) is more suitable for counterfactual explanations (better attainability), (b) yields more robust variable importance explanations, and (c) suggests rejecting the explanations, as the model lacks data or training (state of ignorance).}\label{figure:dataset_toy}
\end{figure}

\subsection*{A summary of our approach}\label{section:summary}

In essence, one might interpret the model learned from the observed data as a compact representation of the sub-population (of the entire population) of which the observed data are representative. Therefore, if the model is (epistemically) uncertain about its prediction on a particular instance due to a lack of observed data, the explanations provided by XAI methods might be irrelevant. This is because such explanations essentially explain the model, whose relevance is higher for input space regions rich in observed data. Therefore, we propose a reject option for explanations based on the degree of epistemic uncertainty, which rejects explanations of those instances that are assigned high degrees of epistemic uncertainty.     

For instances with lower epistemic uncertainty and high degrees of aleatoric uncertainty, it is reasonable to expect that they likely belong to dense regions of the input space where classes are mixed. As a sequel, one might further expect that the model may not be very robust in such a region, \emph{i.e.}, small perturbations in the features of the instances may lead to a change in the model's predictions. In such cases, one may need to use several features to construct informative feature-importance explanations. Such long and ambiguous explanations may challenge human cognitive ability, especially when the total number of features is large, which may be unavoidable in different applications. In contrast, by definition, counterfactual explanations may be close(er) in such cases and appear to be a more promising type of explanation. Therefore, we propose to present counterfactual explanations to the end-user in the case of high aleatoric uncertainty. 

\begin{figure}
\centering
\begin{tikzpicture}
\node (inst) at (0,0) {$x$};
\node[text width=3.5cm, text centered] (rej) at (-2,-2) {\small Situation (c) in Fig~\ref{figure:dataset_toy} \\ insufficient training} ;
\draw[->] (inst) -- (rej) node[near start,left] {\small $EU(x) \nearrow$};
\node[text width=3.5cm, text centered] (AU_check) at (2,-2) {\small Situations (a,b) in Fig~\ref{figure:dataset_toy} \\ check $AU$} ;
\draw[->] (inst) -- (AU_check) node[near start,right] {\small \; $EU(x) \searrow$};
\node[text width=3.5cm, text centered] (AU_high) at (0.5,-4) {\small Situation (a)  \\ counterfactual} ;
\node[text width=3.5cm, text centered] (AU_low) at (3.5,-4) {\small Situation (b)  \\ feature importance} ;
\draw[->] (AU_check) -- (AU_low) node[near start,right] {\small \;$AU(x) \searrow$};
\draw[->] (AU_check) -- (AU_high) node[near start,left] {\small $AU(x) \nearrow$};
\end{tikzpicture}
\caption{Summary of the proposed method.}
\label{figure:method}
\end{figure}
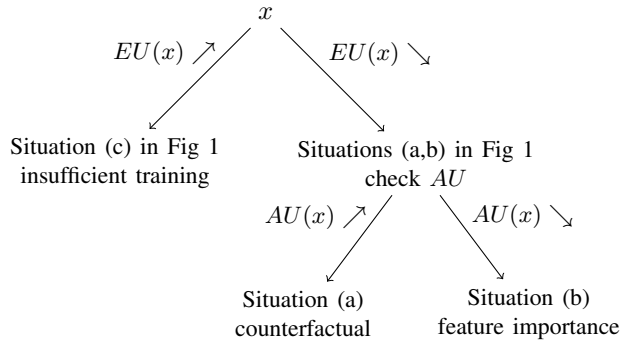

On instances with low degrees of epistemic and aleatoric uncertainty, and given the definition of aleatoric uncertainty, 
one might expect the instance to belong to rather dense region of the input space where classes are (well) separated (by the model). As a sequel, one might further expect the model to be very robust in such a region. In contrast with the previous case, in such regions one could have a chance to construct short(er) and more robust feature-importance explanations. Therefore, we propose to present feature-importance explanations to the end-user in the case of low aleatoric uncertainty. 

Our proposal is illustrated in Figure~\ref{figure:dataset_toy}, and the associated protocol in Figure~\ref{figure:method}. This simple protocol leverages a robust framework to support the field of XAI. Our empirical findings can be summarized as follows:
\begin{itemize}
    \item Significantly negative Spearman’s rank correlations of degrees of aleatoric uncertainty (\emph{i.e.}, lower estimates of the ground-truth aleatoric uncertainty) vs. dissimilarity of counterfactual explanations (\emph{cf.} Section \ref{sec:Counterfactual explanations vs aleatoric uncertainty}) confirm the robustness of counterfactual explanations in cases of high aleatoric uncertainty.
    \item Significantly positive Spearman’s rank correlations of degrees of aleatoric uncertainty vs. robustness of feature-important explanations (\emph{\emph{cf.}} Section \ref{sec: Feature-importance explanations and total uncertainty}) confirm the robustness of feature-important explanations in cases of low aleatoric uncertainty.
    \item Empirical rejections based on high epistemic uncertainty (\emph{cf.} Section \ref{section:experiment_ep}) confirm the usefulness of our proposed reject option. 
\end{itemize}

The rest of the paper is structured as follows: 
Section~\ref{section:background} presents the uncertainty quantification methods used in this study, along with the examined explainability techniques and robustness metrics. Experiments are conducted in Section~\ref{section:experiments}, first demonstrating the motivation and the robustness of the proposed strategy in classical machine learning, followed by its application in a deep learning image classification scenario. Finally, Section~\ref{section:conclusion} concludes the article.

\section{Robust explanations through uncertainty decomposition}\label{section:background}

Let $\mathcal{X}:=\mathcal{X}^1 \times \ldots \times \mathcal{X}^Q := \mathbb{R}^Q$ denote the feature space, and let $\mathbf{X} = \{X^1, \ldots, X^Q\}$ be a finite set of features. Let $Y$ be a class variable, whose possible outcomes are $\mathcal{Y} :=\{y^1, \ldots, y^C\}$. Given a training dataset $\mathbf{D}_{\mathrm{train}} = \{(\vec{x}_n, y_n)\;|\; 1 \leq n \leq N\} \subset \mathcal{X} \times \mathcal{Y}$, the goal of multi-class classification is to learn a classifier
\begin{equation}\label{eq:classifier}
    h: \mathcal{X} \longmapsto \mathcal{Y} \,,
\end{equation}
which predicts for each instance $\vec{x} \in \mathcal{X}$, an outcome $h(\vec{x}) \in \mathcal{Y}$. For scoring models such as deep networks, we denote by $f$ its score output vector within $\mathbb{R}^C$, and $h$ is then obtained by taking $\arg\max f$.


\subsection{Explanations}\label{section:xai}
In this work, we focus on the application of local post-hoc XAI methods (a posteriori explainability) to interpret the predictions of a machine learning model. These methods are divided into two categories: agnostic approaches (model-independent) and formally specific approaches, intrinsic to the architecture, such as methods based on symbolic implementations \citep{Audemardetal22, ShihCD18, IgnatievNM19}. We emphasize agnostic methods, particularly feature attribution techniques, which assign an importance score to each variable for a given prediction: we recall that SHAP \citep{Lundberg17} (Shapley Additive Explanations), is based on game theory \citep{shapley} for a fair distribution of feature impact. Another approach, LIME \citep{Lime16} (Local Interpretable Model-agnostic Explanations), locally approximates the model with an interpretable linear substitute. Counterfactual explanations \citep{Dhurandhar2018} identify the minimal changes needed to alter a prediction (\emph{e.g.}, ``If your income were 10\% higher, your loan would be approved''). Although these methods are scalable and easy to deploy, their reliability may be limited in critical domains (healthcare, justice) due to approximations or biases \citep{IgnatievNM19, Joaogame}. They nevertheless remain popular for their flexibility and applicability to any black-box model. Current approaches focus solely on deriving explanations for ML model predictions. However, as \citep{Miller19} demonstrates, providing explanations without quantifying their utility is insufficient for adoption by ML system users. The quality and usefulness of an explanation ultimately depend on its ability to meet users' needs and expectations. The work of \citep{LaugelECML} explores the risk of generating unjustified post-hoc explanations. In previous work \citep{Audemard22ijcai}, user preferences were incorporated to improve explanation quality, but this doesn't guarantee alignment with user expectations. We aim to bridge this gap by evaluating the robustness of explanations and relating it to the uncertainty involved in their derivation. To achieve this, we propose evaluating both epistemic and aleatoric uncertainties, and the local robustness of explanations relative to data instance $\vec{x}$. On another topic, the authors in \citep{Whojiang2022} show that users' epistemic uncertainty affects their understanding of explanations and XAI methods. When perceived as uncertain, explanations may hinder user acceptance. The work of \citep{UncertaintyXAIChiaburu2024} emphasizes uncertainty's critical yet often overlooked role in XAI for trust building. We stress that better uncertainty integration in XAI could strengthen user trust and promote more responsible AI application in practice. In summary, our approach complements state-of-the-art methods by providing a framework that integrates uncertainty and robustness to estimate explanation-related risks, informing users about potential limitations while significantly improving explanation transparency and acceptability.

\smallskip
This Section focuses on two types of explanations. Explanations based on feature-attribution method are discussed in Section~\ref{section:feature}, while counterfactual explanations are presented in Section~\ref{section:counter}. Section~\ref{section:robustness} details the selected approach to assessing the robustness and similarity of the explanations.

\subsubsection{Feature-importance explanations}\label{section:feature}

Feature attribution methods quantify the relative contribution of each input variable $X^q \in \mathbf{X}$, $q = 1, \ldots, Q$, to a model's prediction for a given instance $\vec{x}_t$ in the test set $\mathbf{D}_{\mathrm{test}} = \{\vec{x}_t\;|\; 1 \leq t \leq T\}$ of size $T$. Formally, for each instance $\vec{x} \in \mathcal{X}$, they compute an importance vector $\boldsymbol{\phi}(\vec{x}) = (\phi_1(\vec{x}), \cdots, \phi_Q(\vec{x}))$ where $\phi_q$ represents the impact of $X^q$ on the model's prediction 
through principles of \textit{perturbation} or \textit{decomposition} \citep{sundararajan2017axiomatic}. In this work, we concentrate on SHAP \citep{Lundberg17}, however, other methods such as LIME \citep{Lime16} are also widely used.
SHAP uses Shapley values \citep{shapley} from game theory, to compute the importance vector $\phi_q$, $q=1,\ldots, Q$. Denoting $\llbracket Q \rrbracket=\{1,\ldots,Q\}$, the Shapley value is defined as:
\begin{equation}\label{eq:shap}
    \phi_q(\vec{x}) = \sum_{\mathbf{S} \subseteq \llbracket Q \rrbracket \setminus q} \frac{|\mathbf{S}|!(Q-|\mathbf{S}|-1)!}{Q!} \left( g(\mathbf{S} \cup \{q\}) - g(\mathbf{S}) \right) \,.
\end{equation}
where $g$ is some value function of a coalition of elements. While one may plug-in various values within Equation~\eqref{eq:shap} (\emph{cf.},~\cite{sundararajan2020many}), the classical SHAP values we will consider here correspond to take $G(\mathbf{S})=\mathbb{E}(f(X)|X_{\mathbf{S}}=x_{\mathbf{s}})$, that is the conditional expectation of the score function for the class, with features within the subset $S$ being fixed.

    

While extensively used, these methods have notable limitations. SHAP assumes feature independence (potentially unrealistic) while LIME is sensitive to local perturbations. Their scores do not imply causality, and they often ignore the instance's predictive uncertainty, which should be quantified while potentially diverging from human expertise, thus requiring cautious interpretation.

\subsubsection{Counterfactual explanations}\label{section:counter}

A contrastive explanation identifies the minimal changes needed for an instance to alter a model's decision. These explanations answer the question: ``what should be modified to obtain a different result?''. In the literature, these explanations can take several forms. Recent work by~\citep{Dhurandhar2018,Miller2021Contrastive, Dhurandhar2024} proposes interesting approaches, but presents a major limitation: their contrastive explanations, while mathematically valid, do not necessarily correspond to realistic instances from the training set, which limits their practical utility.

For our study, we therefore favor a more concrete approach. In order to explain the prediction for an instance $\vec{x}_t\in \mathbf{D}_{\mathrm{test}}$, we identify its nearest neighbor $\Tilde{\vec{x}_t}$ in the training set $\mathbf{D}_{\mathrm{train}}$ 
with a different label $y^c \neq h(\vec{x}_t) \in \mathcal{Y}$. 
A counterfactual explanation, denoted by $\Tilde{\vec{x}_t}$, is defined as follows:
\begin{equation}\label{eq:counterfactuals}
    \Tilde{\vec{x}_t} = \underset{{\Tilde{\vec{x}},\; h(\vec{x}_t) \neq y^c}}{\arg\min}||\vec{x}_t - \Tilde{\vec{x}}||_2.
\end{equation}
This method guarantees that the proposed counterfactual explanation is always a realistic and interpretable instance.

\subsection{Explanation robustness}\label{section:robustness} 
To provide insights into the robustness of the explanations, we consider two different metrics that assess robustness against noisy executions~\citep{Jiang2024, Wei2023}.
Lipschitz continuity is extended to the case of explanations by feature-importance in~\citep{alvarezmelis2018}. Let $\phi$ be a mapping from the input space to the set of possible explanations, where for each instance $\vec{x}_t$, $\phi(\vec{x})$ is a feature-importance vector. Let $||\cdot||_2$ be the $L^2$-norm. The discrete local Lipschitz continuity, for any instance $\vec{x}_t\in \mathbf{D}_{\mathrm{test}}$, is defined as follows:
\begin{equation}\label{eq:robustness_level}
    L(\vec{x}_t) = \max_{\vec{x}\in\mathcal{N}(\vec{x}_t)}\frac{||\phi(\vec{x}_t) - \phi(\vec{x})||_2}{||\vec{x}_t - \vec{x}||_2} \,,
\end{equation}
where $\mathcal{N}(\vec{x}_t)$ is the set of elements contained within the ball of radius $\epsilon$ centered at $\vec{x}_t$, defined as follows:
\begin{equation}
    \mathcal{N}(\vec{x}_t) = \{\vec{x} \in \mathcal{X}\;|\; ||\vec{x}_t - \vec{x}||_2 \leq \epsilon\}\,.
\end{equation}
For the experiments, the authors suggest using $\epsilon = 0.1$ and estimating $L(\vec{x}_t)$ using 30 samples $\vec{x}$ within $\mathcal{N}(\vec{x}_t)$ (\emph{cf.},~\citep{alvarezmelis2018}).

In the case of a counterfactual explanation, it must be attainable by definition (\emph{c.f.}~\citep{guidotti24}). A measure of similarity~\citep{Pawelczyk2020} between the predicted instance $\vec{x}_t \in \mathcal{X}$ and its counterfactual $\Tilde{\vec{x}}_t$ is the latent distance $d$, defined as follows:
\begin{equation}
    d(\vec{x}_t, \Tilde{\vec{x}}_t) = ||\vec{z}_t - \Tilde{\vec{z}}_t||_p \,,
\end{equation}
where $||\cdot||_p$ is the $L^p$-norm. Certain features may be protected for attainability reasons, leading to the mapping of $\vec{x}_t$ to a latent space $\vec{z}_t$. To simplify calculations, we assume in our experiments that all variables are unrestricted, \emph{i.e.}, $\vec{x}_t = \vec{z}_t$. For the distance measure, we consider all variables to be equally weighted and use a simple Euclidean norm $d$, given by:
\begin{equation}\label{eq:attaignability}
    d(\vec{x}_t, \Tilde{\vec{x}}_t) = ||\vec{x}_t - \Tilde{\vec{x}}_t||_2.
\end{equation}
This measure characterizes how difficult it is to transform the predicted observation into its counterfactual. The goal of this article is to demonstrate that certain explanations are more relevant in specific contexts and that model uncertainty can be highly useful, sometimes even serving as an explanation itself.

\subsection{Uncertainty quantification}\label{section:uq}

In the literature, two primary types of uncertainty are typically distinguished~\citep{Hora1996}: aleatoric uncertainty (data uncertainty) and epistemic uncertainty (model uncertainty). Aleatoric uncertainty arises from the stochastic nature of the data-generating process, while epistemic uncertainty is associated with a lack of knowledge. The former is generally considered irreducible, whereas the latter can be mitigated by acquiring additional data. Figure~\ref{figure:dataset_toy} presents these two types of uncertainty on a toy dataset.

To disentangle aleatoric and epistemic uncertainty, numerous methods have been proposed in recent years~\citep{huang2023}. These methods can be grouped into four main families of models.
Bayesian methods compute a posterior distribution over a model’s parameters and construct a second-order distribution over class probabilities via Bayesian model averaging~\citep{kendall2017, Depeweg2018}.
Evidential deep learning adopts a frequentist perspective and aims to minimize both model error and epistemic uncertainty, which is represented by a second-order probability. For multi-class classification problems, the Dirichlet distribution is typically used~\citep{Sensoy2018, Malinin2018, Malinin2019, Ulmer2023}.
Ensemble methods capture epistemic uncertainty through ensemble diversity. This can be achieved by randomizing training data via bootstrapping, varying neural network architectures using techniques such as MC Dropout, or randomizing the optimizer as in Deep Ensembles~\citep{Lakshminarayanan2017, nguyen2023learning, Mobiny2021, Shaker2020}.
Density-based and distance-based methods typically model aleatoric and epistemic uncertainty directly~\citep{Nguyen2022, Hoarau2024-ml, van-amersfoort20a, sun2022, Liu2020}. However, these methods can also construct a second-order distribution in a post-processing step~\citep{Charpentier2020, Charpentier2022}.

Our main study investigates three state-of-the-art approaches to disentangle aleatoric and epistemic uncertainty. 
First, we used an entropy-based uncertainty estimation, following the entropy decomposition method described in~\citep{Shaker2020}. Here, the posterior distribution is approximated by a finite ensemble of estimators (a random forest here). If $p_k$ is the predicted probability of the k$^{th}$ model, the aleatoric uncertainty $\mathrm{AU}_{ens}$ in $\vec{x}_t$ can be obtained as follows:
\begin{equation}\label{eq:au_entropy}
    \mathrm{AU}_{ens}({\vec{x}_t}) = - \frac{1}{K}\sum_{k=1}^K\sum_{y\in\mathcal{Y}}p_k(y| \vec{x}_t)\log_2p_k(y| \vec{x}_t) \,,
\end{equation}
where $K$ is the number of ensemble members.
Total uncertainty $\mathrm{TU}_{ens}$ can be approximated in terms of Shannon's entropy of the predictive posterior distribution as follows:
\begin{equation}
    \begin{split}
    \mathrm{TU}_{ens}(\vec{x}_t) = -\sum_{y\in\mathcal{Y}}&\left(\frac{1}{K}\sum_{k=1}^K p_k(y|\vec{x}_t)\right)\\
    &\times\log_2\left(\frac{1}{K}\sum_{k=1}^K p_k(y| \vec{x}_t)\right).
    \end{split}
\end{equation}
Finally, the epistemic uncertainty $\mathrm{EU}_{ens}$ can be derived through additive decomposition as $\mathrm{EU}_{ens}(\vec{x}_t) = \mathrm{TU}_{ens}(\vec{x}_t) - \mathrm{AU}_{ens}(\vec{x}_t)$.

We also investigate a centroid-based approach presented in~\citep{van-amersfoort20a}. To compute the model's uncertainty, the authors proposed measuring the distance from the incoming observation to the class centroids using a radial basis function:
\begin{equation}
    \mathrm{U}_y(\vec{x}_t, \textbf{e}_y) = \exp\left[-\frac{\frac{1}{I}||\textbf{W}_y f(\vec{x}_t) - \textbf{e}_y||^2_2}{2\sigma^2}\right] \,,
\end{equation}
with $h$ the model, $\textbf{e}_y$ the centroid associated with class $y$ and $\textbf{W}_y$ a weight matrix of
size $I$ (centroid size) by $J$ (feature extractor output size) in the case of a deep architecture. The length scale $\sigma$ is an hyper-parameter of the method.
The epistemic certainty $\mathrm{EC}$ associated with instance $\vec{x}_t$ is then obtained by measuring the distance from the incoming observation to the nearest class centroid:
\begin{equation}\label{eq:eu_centroid}
    \mathrm{EC}(\vec{x}_t) = \max_{y\in\mathcal{Y}}\mathrm{U}_y(\vec{x}_t, \textbf{e}_y) \,.
\end{equation}
One can then obtain $\mathrm{EU}_{cen}(\vec{x}_t) = 1 / \mathrm{EC}(\vec{x}_t)$ while $\mathrm{AU}_{cen}$ is computed based on the relative distances between all clusters (entropy of the obtained softmax distribution).

To locally estimate the epistemic uncertainty of the model, the authors in~\citep{Hoarau2024-ml} propose using Evidential K-NN~\citep{denoeux1995}, which is based on the mathematical theory of evidence~\citep{Dempster1967,shafer1976}. Given the $K$ nearest neighbors of $\vec{x}_t$ according to a distance metric $d$, the model outputs a belief function $m_t$~\citep{shafer1976,denoeux1995} defined over $2^\mathcal{Y}$ as the prediction for $\vec{x}_t$, such that $m(A) \in [0,1]$ and $\sum_{A \subseteq \mathcal{Y}}m(A)=1$. The epistemic uncertainty $\mathrm{EU}_{bel}$ is then obtained from the Non-Specificity~\citep{dubois1987} as follows:
\begin{equation}\label{eq:nonspe}
    \mathrm{EU}_{bel}(\vec{x}_t) = \sum_{A\subseteq \mathcal{Y}} m_t(A)\text{log}_2(|A|)\,,
\end{equation}
and the aleatoric uncertainty $\mathrm{AU}_{bel}$ is obtained as follows~\citep{klir1990uncertainty}:
\begin{equation}\label{eq:discord}
    \mathrm{AU}_{bel}(\vec{x}_t) = - \sum_{A\subseteq \mathcal{Y}} m_t(A)\text{log}_2({\mathrm{BetP}(A, m_t)})\,,
\end{equation}
where $\mathrm{BetP}$ is the probability measure associated with the following probability mass:
\begin{equation}
    \mathrm{BetP}(\{y\}, m_t) = \sum_{A\subseteq \mathcal{Y}, \;\;y\in A}\frac{m_t(A)}{|A|}\,.
\end{equation}
Aleatoric uncertainty is therefore higher in dense, conflicting regions, while epistemic uncertainty is higher in low-density regions. Note that $\mathrm{BetP}$ actually corresponds to applying a Laplace principle of indifference for each subset having positive mass $m_t(A)$.

We include in the appendix another approach using a relative-likelihood based strategy elaborated in~\citep{Nguyen2022,senge2014}, 
as this approach is only defined for the binary case. In a few words, within this approach two highly plausible classes result in high epistemic uncertainty while two implausible classes reflect high aleatoric uncertainty.

\section{Experiments}\label{section:experiments}

To ensure a comprehensive evaluation of our protocol, we present different sets of experiments on $10$ tabular datasets from the UCI repository~\citep{Dua2019} with varied numbers of classes/features/instances, and the MNIST image dataset, that examine the relationship between explanation robustness and uncertainty quantification. Our experiments aims at showing that uncertainty quantification can serve as a guide to relevant explanation, and sometimes as an explanation itself.

A description of the datasets is given in Table \ref{tab:Datasets} of Appendix \ref{appendix:datasets}. The source code used in this paper is submitted as supplementary material and shall be made public upon the acceptance of this paper.

Section~\ref{section:experiment_rob} focuses on the robustness of explanations, showing that aleatoric uncertainty can be used to select more appropriate explanations. Section~\ref{section:experiment_ep} treats high epistemic uncertainty as a form of explanation by analyzing the most ignorant predictions and arguing that explanations for out-of-distribution instances lack support and meaning. This behavior is observed in both classical machine learning and deep learning classification tasks.

\subsection{Robustness of explanations}\label{section:experiment_rob}

\begin{figure}[b]
    \centering
    \begin{subfigure}[t]{0.24\linewidth}
    \includegraphics[width=\linewidth]{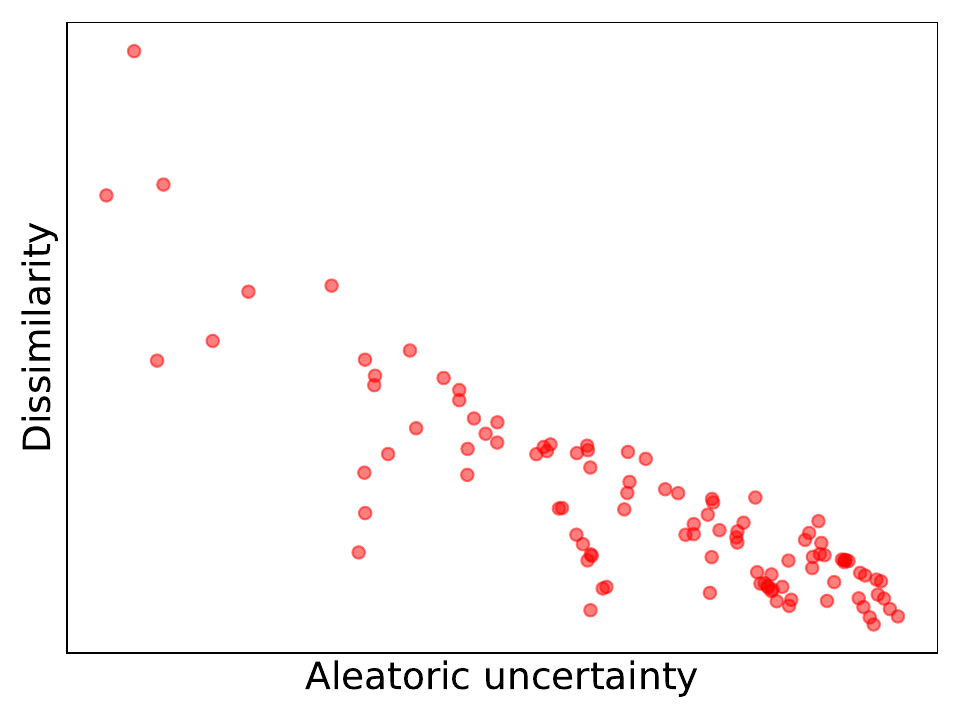}
    \caption{Liver}
    \end{subfigure}
    \begin{subfigure}[t]{0.24\linewidth}
    \includegraphics[width=\linewidth]{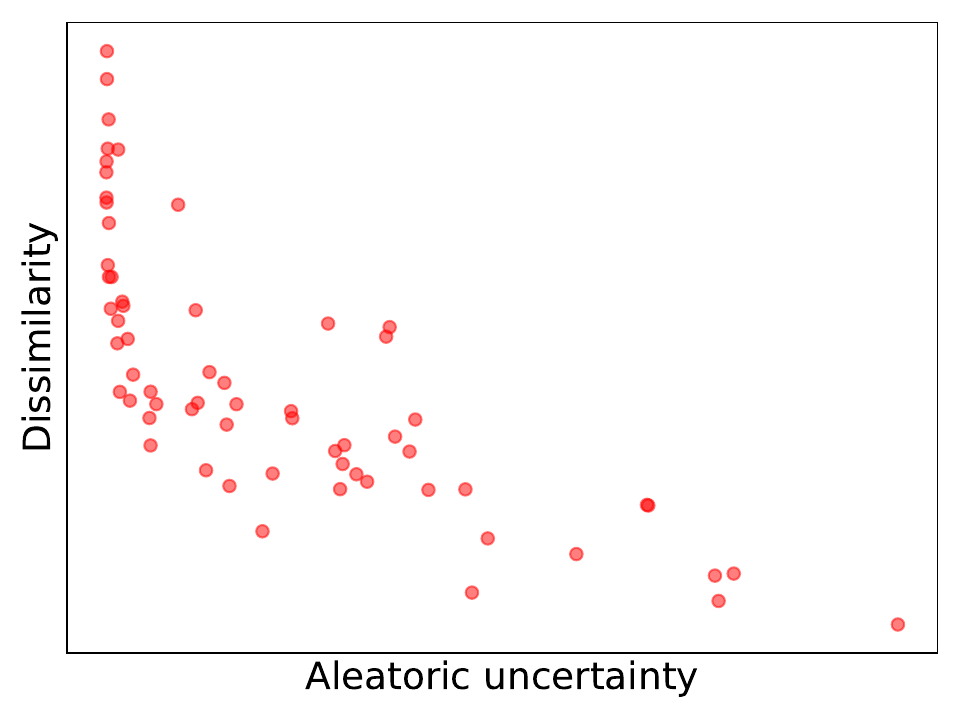}
    \caption{Sonar}
    \end{subfigure}
    \begin{subfigure}[t]{0.24\linewidth}
    \includegraphics[width=\linewidth]{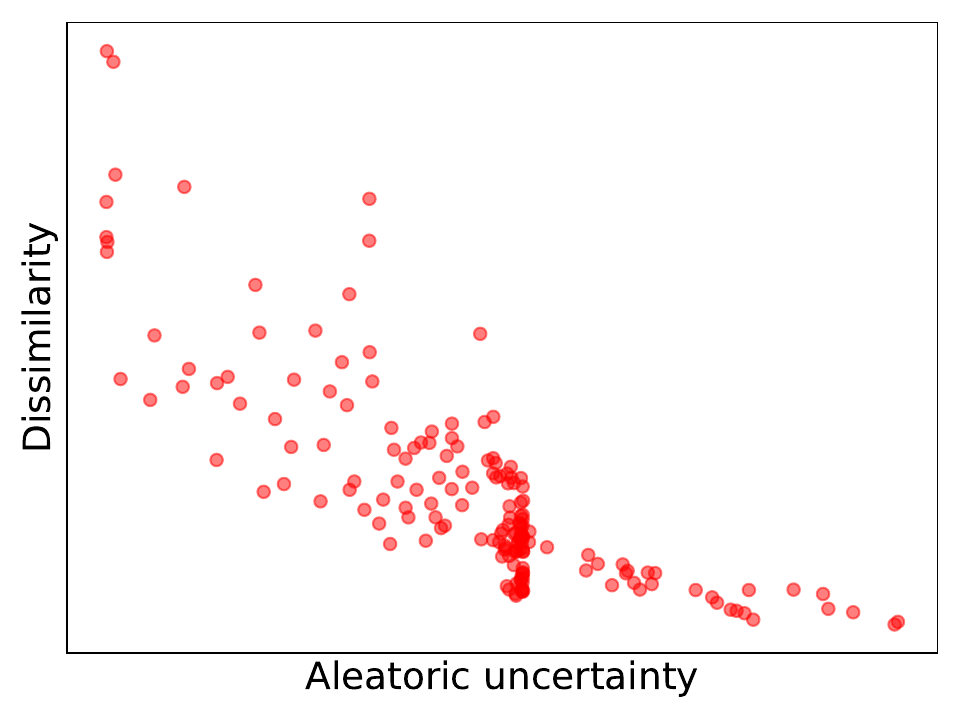}
    \caption{Breast Cancer}
    \end{subfigure}
    \begin{subfigure}[t]{0.24\linewidth}
    \includegraphics[width=\linewidth]{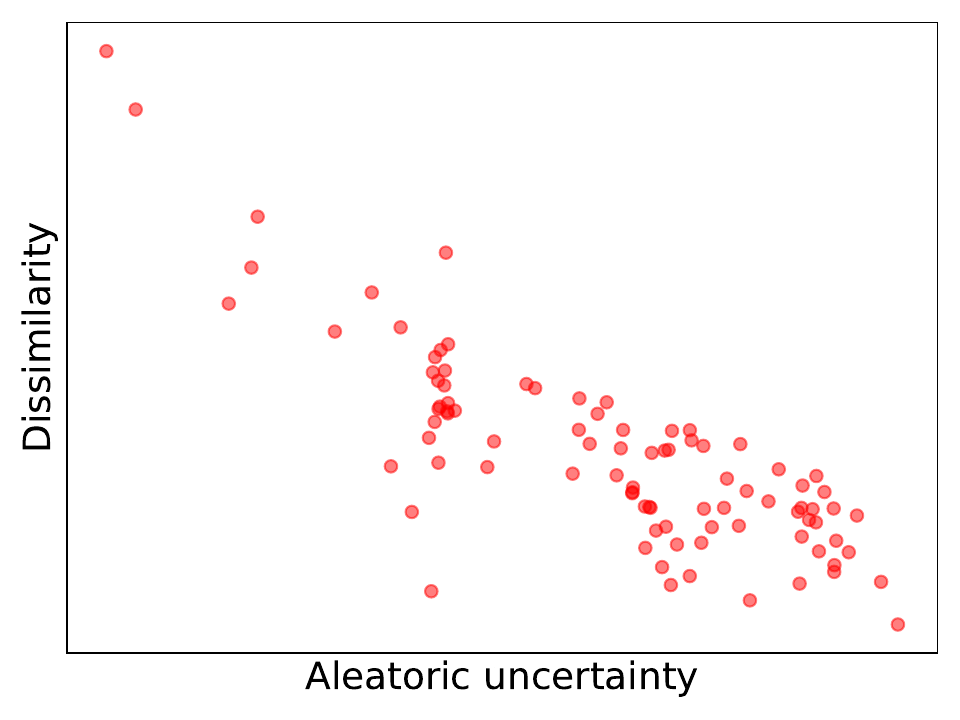}
    \caption{Heart}
    \end{subfigure}
    \caption{Aleatoric uncertainty \emph{vs.} dissimilarity, leveraging a correlation between similarity and uncertainty. More aleatoric uncertainty leads to more similar counterfactual explanations.}
    \label{fig:ce_robust_uncertainty}
\end{figure}

We assess here the robustness of two types of explanations: feature attribution-based (using SHAP) and counterfactual explanations. 

For each dataset, we analyze the correlation between uncertainty and robustness, as defined in Equations~\eqref{eq:robustness_level} and~\eqref{eq:attaignability}. To be readable, figures are extracted from a random run and are constructed with the uncertainty decomposition described in Equations~\eqref{eq:nonspe} and~\eqref{eq:discord}. However, Spearman's rank correlation are averaged over 100 runs and concern all decomposition methods, showing their overall consistency. More results are provided in Appendix~\ref{appendix:counterfactuals} and~\ref{appendix:feature-att}, along with the detailed 
experimental protocol.

\subsubsection{Counterfactual explanations} \label{sec:Counterfactual explanations vs aleatoric uncertainty} 

Let us start with attainability of counterfactuals, showing that aleatoric uncertainty is a good indicator of regions where counterfactuals are stable and can therefore be used as reliable explanations. 


Figure~\ref{fig:ce_robust_uncertainty} illustrates the relationship between aleatoric uncertainty and the (un)robustness of counterfactual explanations across four datasets. In this context, samples with higher uncertainty tend to produce counterfactual explanations that remain notably consistent, indicating greater attainability and hopefully a higher user acceptability. 


\begin{table}
\caption{Spearman's rank correlation of aleatoric uncertainty for each studied method \emph{vs.} dissimilarity of counterfactual explanations, exhibiting a negative correlation.}
\label{tab:ce_robust_uncertainty}
\begin{center}
\begin{small}
\begin{sc}
\begin{tabular}{@{}lccc@{}}
\toprule
Dataset & \multicolumn{3}{c}{Correlations} \\
        & $AU_{bel}$ & $AU_{cen}$ & $AU_{ens}$\\
\midrule
Breast Cancer \;& -0.83 & -0.52 & -0.29\\
Ecoli   & -0.53 & -0.21 & -0.13\\
Glass   & -0.55 & -0.27 & 0.15\\
Heart   & -0.79 & -0.33 & -0.13\\
Ionosphere  & -0.61 & -0.04 & 0.30\\
Iris    & -0.58 & -0.73 & -0.50\\
Liver   & -0.82 & -0.38 & -0.11\\
Park.   & -0.89 & -0.66 & -0.69\\
Sonar   & -0.82 & -0.32 & -0.07\\
Wine    & -0.45 & -0.49 & -0.29\\
\bottomrule
\end{tabular}
\end{sc}
\end{small}
\end{center}
\end{table}

Table~\ref{tab:ce_robust_uncertainty} further confirms the negative correlation between uncertainty and explanation dissimilarity across all 10 datasets, for almost all decomposition methods. Note that $AU_{bel}$ and $AU_{cen}$ provide higher correlation values in general, indicating that more local uncertainty quantification should be privileged to better select counterfactual. Appendix~\ref{appendix:counterfactuals} provides results completed with statistical significance information, indicating all significant correlations, confirming that aleatoric uncertainty quantification is a reliable indicator of counterfactual stability.

\begin{figure}[b]
    \centering
    \begin{subfigure}[t]{0.24\linewidth}
    \includegraphics[width=\linewidth]{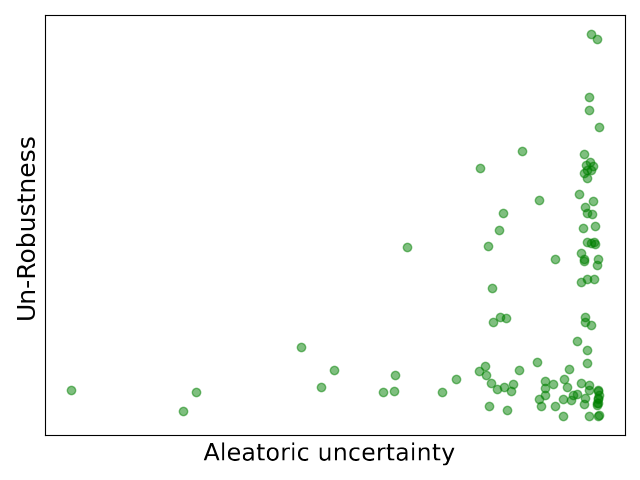}
    \caption{Liver}
    \end{subfigure}
    \begin{subfigure}[t]{0.24\linewidth}
    \includegraphics[width=\linewidth]{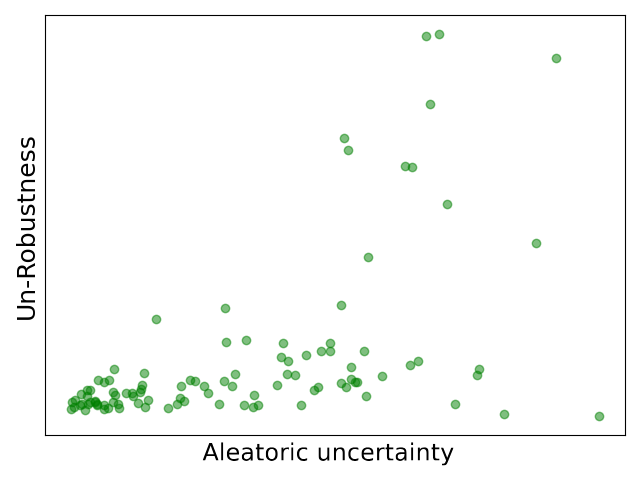}
    \caption{Ecoli}
    \end{subfigure}
    \begin{subfigure}[t]{0.24\linewidth}
    \includegraphics[width=\linewidth]{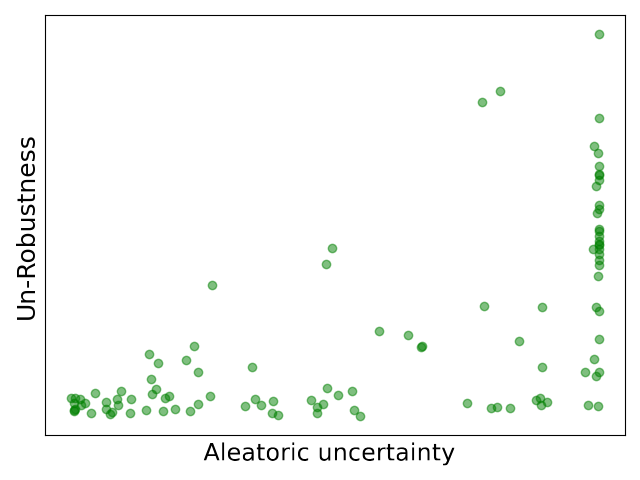}
    \caption{Ionosphere}
    \end{subfigure}
    \begin{subfigure}[t]{0.24\linewidth}
    \includegraphics[width=\linewidth]{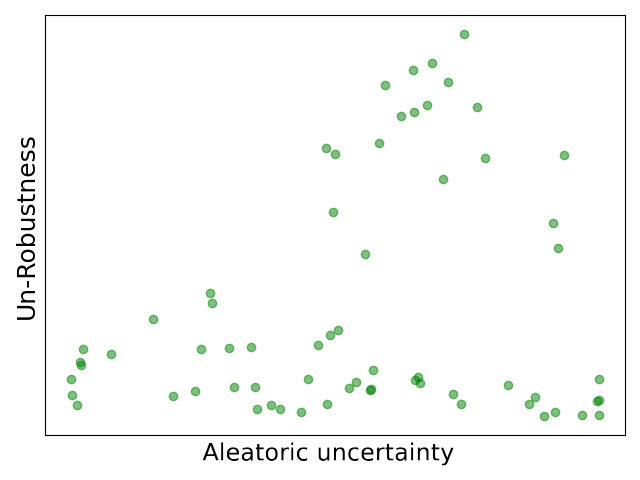}
    \caption{Glass}
    \end{subfigure}
    \caption{Aleatoric uncertainty \emph{vs.} un-robustness, leveraging a correlation between un-robustness and uncertainty. More aleatoric uncertainty leads to less robust SHAP explanations.}
    \label{fig:shap_robust_uncertainty}
\end{figure}

\subsubsection{Feature-importance explanations}  \label{sec: Feature-importance explanations and total uncertainty}

This section focuses on the robustness of feature-importance explanations. Our aim is to identify regions where these explanations are more robust, highlighting that aleatoric uncertainty can also serve as a useful indicator to know when such kinds of explanations are relevant and trustworthy. Epistemic uncertainty reflects how precisely one can estimate true aleatoric uncertainty~\citep{eyke2019}, and high epistemic uncertainty may be the sign of a high/low aleatoric uncertainty. 
Therefore, in this experiment, we consider aleatoric uncertainty both without rejection, and with rejecting 30\% of instances with highest epistemic uncertainty (\emph{cf.}, Section~\ref{section:experiment_ep}), to better capture the full extent of true aleatoric uncertainty (this approach was not used in the previous section, as counterfactuals are more attainable in dense regions with high aleatoric uncertainty, where epistemic uncertainty is low in all cases).

Figure~\ref{fig:shap_robust_uncertainty} shows the relationship between aleatoric uncertainty and the un-robustness of SHAP explanations on four datasets. Samples with lower uncertainty tend to yield more stable SHAP explanations, reflecting higher robustness. This observed consistency enhances the reliability of the explanations and highlights the need for robust explanation protocols in the presence of uncertainty.


\begin{table}
\caption{Spearman's rank correlation of aleatoric uncertainty for each studied method \emph{vs.} robustness of feature-importance explanations, with the full test data and with rejecting 30 \% of highly epistemic uncertain instance. }
\label{tab:shap_robust_uncertainty}
\begin{center}
\begin{small}
\begin{sc}
\begin{tabular}{@{}lcccccc@{}}
\toprule
Dataset & \multicolumn{6}{c}{Correlations} \\
        & \multicolumn{2}{c}{$AU_{bel}$} & \multicolumn{2}{c}{$AU_{cen}$} & \multicolumn{2}{c}{$AU_{ens}$}\\
        & Full & 30\% & Full & 30\% & Full & 30\% \\
\midrule
B. Cancer \;& 0.13 & 0.43 & 0.28 & 0.37 & 0.36 & 0.07 \\
Ecoli   & 0.56 & 0.57 & 0.08 & 0.12 & 0.37 &  0.39 \\
Glass   & 0.36 & 0.53 & 0.19 & 0.33 & 0.14 & 0.11 \\
Heart   & 0.21 & 0.48 & 0.29 & 0.43 & 0.37 & 0.44 \\
Iono.  & -0.30 & 0.18 & 0.05 & 0.11 & 0.15 & 0.30 \\
Iris    & 0.68 &0.68 & 0.33 & 0.26 & 0.56 & 0.17 \\
Liver   & 0.41 & 0.50 & 0.12 & 0.14 & 0.23 & 0.32 \\
Park.   & 0.46 & 0.46 & 0.26 & 0.06 & 0.41 & 0.32 \\
Sonar   & -0.15 & -0.05 & 0.05 & 0.19 & 0.22 & 0.17 \\
Wine    & -0.31 & 0.52 & 0.38 & 0.38 & 0.39& 0.39 \\
\bottomrule
\end{tabular}
\end{sc}
\end{small}
\end{center}
\end{table}

The Spearman's rank correlation analysis presented in Table~\ref{tab:shap_robust_uncertainty} confirms that this relationship is positive (except for a few cases), emphasizing the potential advantage of employing our robust explanation protocol based on aleatoric uncertainty. This result generally holds for the likelihood-based method examined in this study, as shown in Appendix~\ref{appendix:feature-att}. While not systematic, we see that for almost all datasets, reducing attention to low epistemic data (by rejecting 30\%) increases our correlation, sometimes drastically (from -0.3 to 0.5 for $AU_{bel}$ on Wine). Again, this is mainly true for the density-based methods, suggesting that more local approaches are better to disentangle uncertainty sources and to recommend robust explanations. 


The next section considers in more detail the rejection of explanations associated with epistemically uncertain predictions (such as out-of-distribution instances or lack of training data). we show in particular reject curves that can be used to select the quantity of rejected explanations, as well as display the most epistemically uncertain data, for which provided explanations indeed make little sense. 

\subsection{Epistemic uncertainty rejections}\label{section:experiment_ep}

This section focuses on a new type of explanation. We claim that predictions with high epistemic uncertainty result from either a lack of training or an out-of-distribution test instance. As a result, a reliable explanation based on the most well-known techniques cannot be provided. Instead, the model's epistemic uncertainty forms an explanation in itself: ``This prediction has been made because the model lacks training''. We investigate the rejection of counterfactual and feature-attribution explanations in both classical machine learning and deep learning tasks. The experimental protocol is 
detailed in Appendix~\ref{appendix:rejection}.

\subsubsection{Feature-attribution and Machine learning classification}

\begin{figure*}[t]
    \centering
    \begin{subfigure}[t]{0.24\linewidth}
    \includegraphics[width=\linewidth]{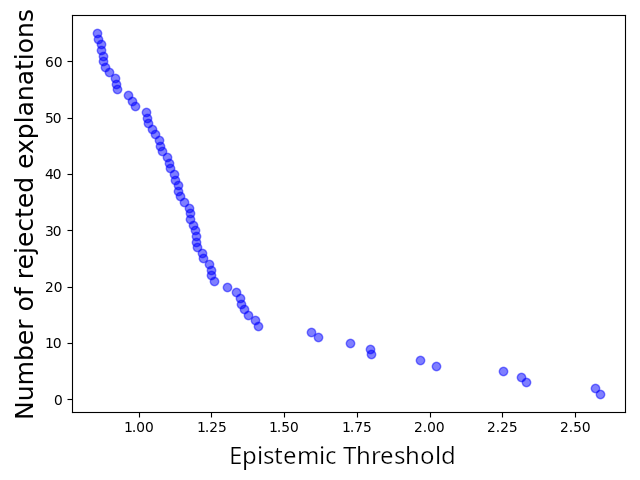}
    \caption{Epistemic \emph{vs.} Rejection}\label{subfig:ep_vs_rej_ml}
    \end{subfigure}
    \begin{subfigure}[t]{0.25\linewidth}
    \includegraphics[width=\linewidth]{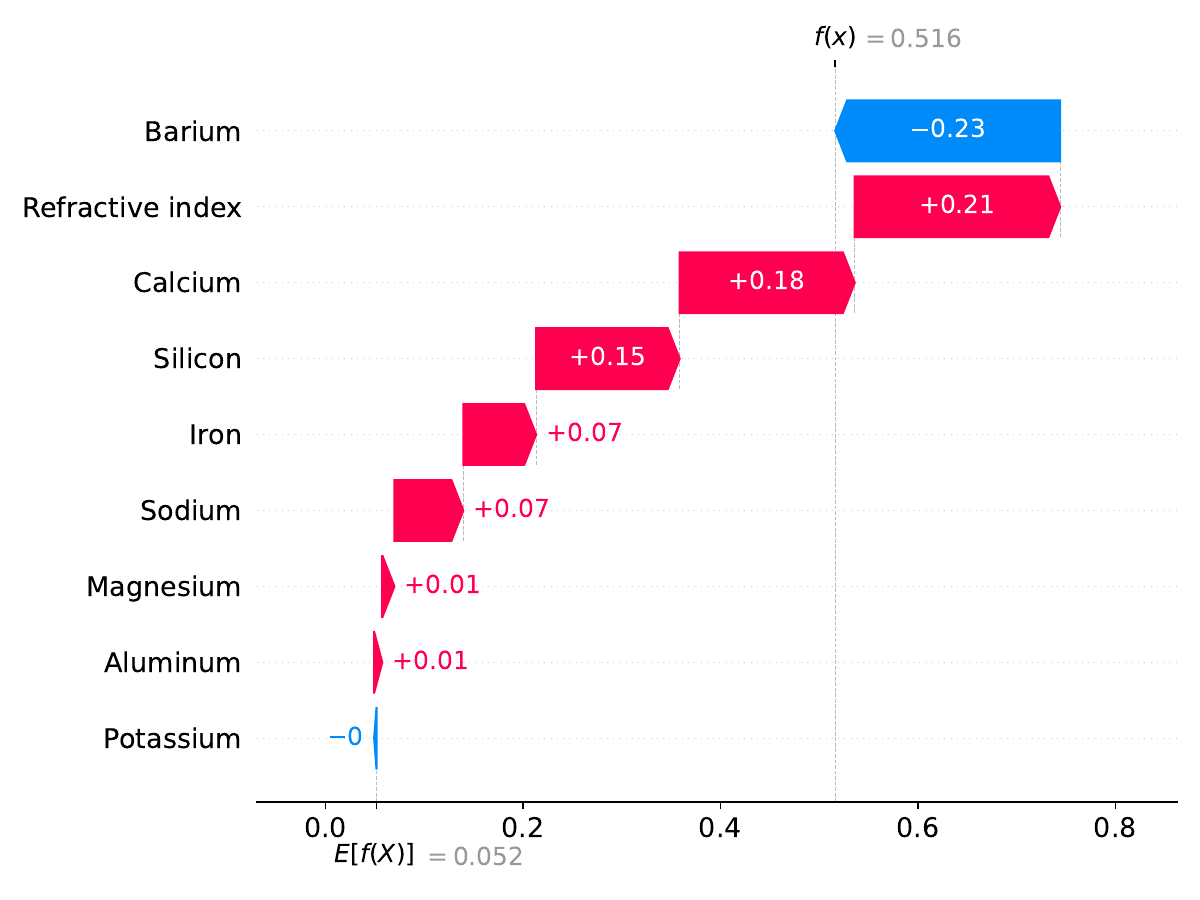}
    \caption{Explanation}\label{subfig:exp_ml}
    \end{subfigure}
    \begin{subfigure}[t]{0.25\linewidth}
    \includegraphics[width=\linewidth]{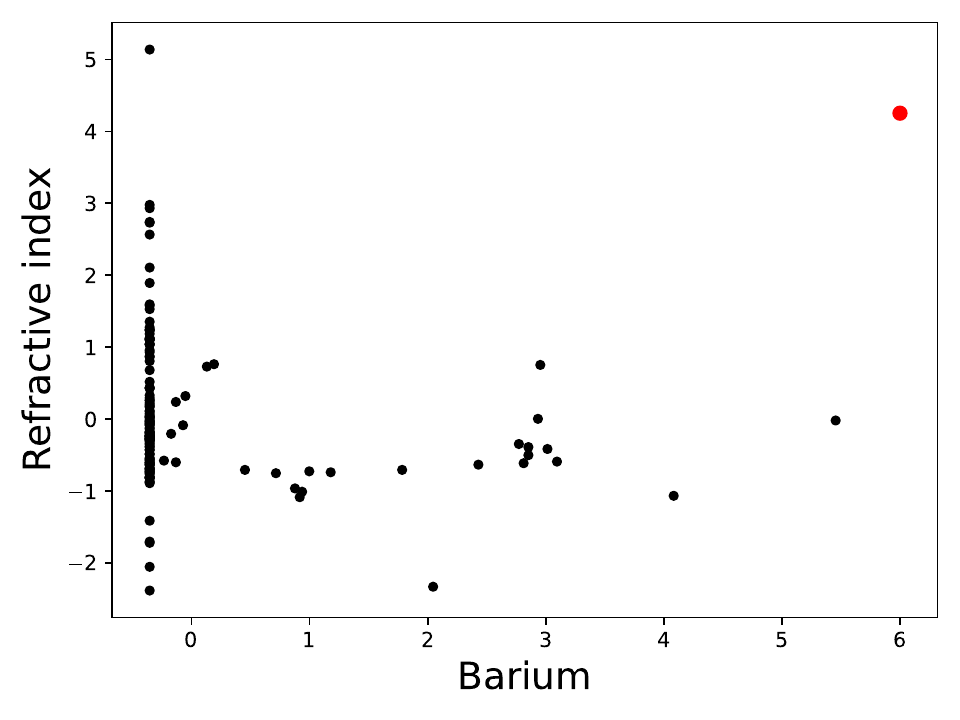}
    \caption{2D representation}\label{subfig:2d_ml}
    \end{subfigure}
    \caption{Rejection based on epistemic uncertainty on Glass dataset: \ref{subfig:ep_vs_rej_ml} Rejection curve based on the epistemic uncertainty thresholds, \ref{subfig:exp_ml} Explanation for the most (epistemic) uncertain test instance, \ref{subfig:2d_ml} Representation on the two most important variables (train set in black and test instance in red, the instance is an outlier). }
    \label{fig:rejection1}
\end{figure*}

Figure~\ref{fig:rejection1} underscores the importance of rejecting feature-attribution explanations for predictions that exhibit high epistemic uncertainty on the Glass dataset, ensuring that only reliable data, or data with low epistemic uncertainty, contributes to the decision-making process.
Figure~\ref{subfig:ep_vs_rej_ml} shows the rejection curve for all test instances based on epistemic uncertainty. 

Naturally, a lower threshold results in a higher number of rejected explanations. To illustrate our point, we consider the test instance associated with the most epistemically uncertain prediction. Its corresponding SHAP explanation is presented in Figure~\ref{subfig:exp_ml}. By considering the two most important features explaining the prediction (the Barium and Refractive Index variables) and plotting the training set on this 2D plane in Figure~\ref{subfig:2d_ml}, we observe that the test instance (in red) is an outlier, and those variables do not effectively represent this instance.
By filtering out high-epistemic-uncertainty samples, our approach yields more stable explanations, ultimately fostering greater trust and interpretability in automated decision-making systems.

\subsubsection{Counterfactuals and deep learning image classification}\label{section:experiment_deep}

\begin{figure*}[t]
    \centering
    \begin{subfigure}[t]{0.24\linewidth}
    \includegraphics[width=\linewidth]{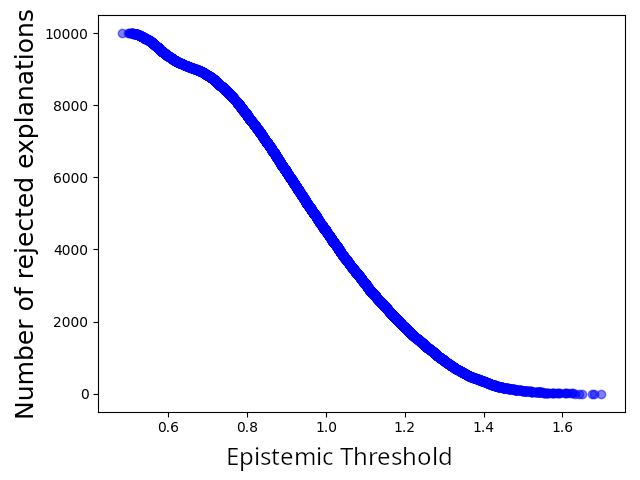}
    \caption{Epistemic \emph{vs.} Rejection}\label{subfig:ep_vs_rej_deep}
    \end{subfigure}
    \begin{subfigure}[t]{0.25\linewidth}
    \includegraphics[width=\linewidth]{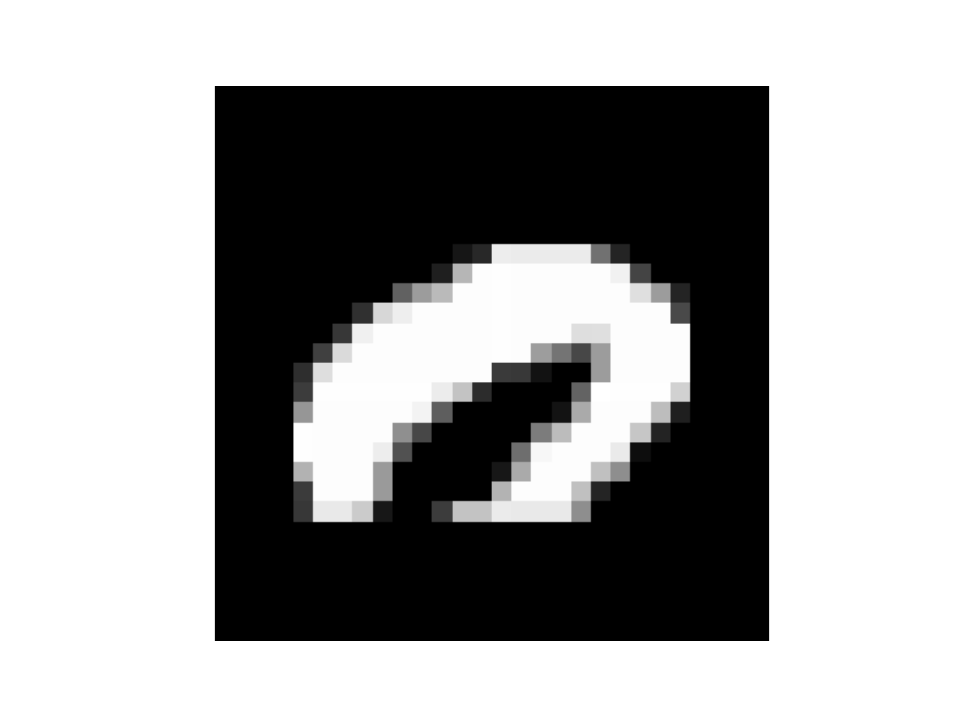}
    \caption{Max. epistemic}\label{subfig:max_deep}
    \end{subfigure}
    \begin{subfigure}[t]{0.25\linewidth}
    \includegraphics[width=\linewidth]{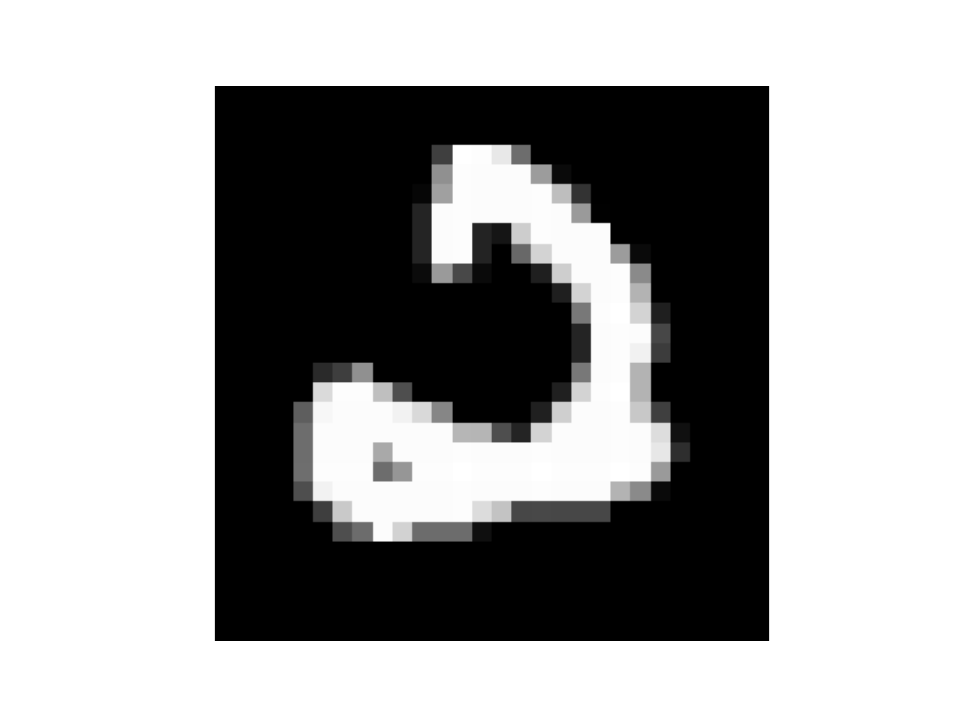}
    \caption{Count. explanation}\label{subfig:exp_deep}
    \end{subfigure}
    \caption{Rejection based on epistemic uncertainty for MNIST dataset: \ref{subfig:ep_vs_rej_deep} Rejection curve based on the epistemic uncertainty thresholds, \ref{subfig:max_deep} Most (epistemic) uncertain test instance, \ref{subfig:exp_deep} Counterfactual explanation (which should clearly be rejected) associated with the test instance.}
    \label{fig:rejection2}
\end{figure*}

Figure~\ref{subfig:ep_vs_rej_deep} presents the rejection curve of counterfactual explanations for all test instances of the MNIST image dataset~\citep{Lecun1998} based on epistemic uncertainty. As previously observed, a lower threshold results in a higher number of rejected explanations.
Again, we consider the test instance associated with the most epistemically uncertain prediction, shown in Figure~\ref{subfig:max_deep}. Its corresponding counterfactual explanation is displayed in Figure~\ref{subfig:exp_deep}. The phenomenon is even more evident in real-world applications and with image data: regardless of the model's prediction for the instance in Figure~\ref{subfig:max_deep}, the generated explanation offers no meaningful insight to the end user and should be rejected.
A counterfactual explanation should make sense within the context of the possible classes, and an out-of-distribution sample cannot lead to a reliable or interpretable explanation.

\section{Conclusion}\label{section:conclusion}

We propose leveraging the complementary nature of uncertainty quantification and notions of explanations to construct guidelines to provide context-dependent, human-friendly explanations, that can then help to enhance their acceptability among users. More precisely, we propose to present counterfactual explanations and feature-importance explanations to the end-user in the cases of high and low aleatoric uncertainty, respectively. We also propose an epistemic uncertainty-based reject option to detect (and reject) non-reliable and/or irrelevant explanations. Multiple experimental studies confirm the usefulness of our proposals in various settings, constituted by the choices of machine learning model, data type, degree of uncertainty, and explanation method.  

By thoroughly revisiting various notions of uncertainty and explanation methods, and elaborating on how to combine them to construct context-dependent, human-friendly explanations, which is largely underexplored, we hope that this paper will facilitate the development of UQ and XAI themselves, and the emerging research direction on seeking their meaningful mixtures in applications. Moreover, we hope that our proposals, which are indeed intuitive, accurate, and compatible with existing/future libraries/proposals for UQ and XAI, will open the path to trustworthier AI.


\bibliography{ref/ref}
\bibliographystyle{style}

\newpage
\appendix

\section{Notations and Acronyms}\label{appendix:Notations}

\begin{table}[http]
\centering
\begin{tabular}{ll}
    \hline
        \textbf{Notations/Acronyms} & \textbf{Description} \\
    \hline
        $Y$ & Class (random) variable \\
        $X^q$ & A feature (random) variable \\
        $\mathbf{X}$ & Feature vector $\{X^1, \dots, X^Q \}$\\
        $\mathcal{X}^q$ & Range of feature $X^q$\\
        $\mathcal{X}$ & Input space $\mathcal{X}^1 \times \ldots \times \mathcal{X}^Q$ \\
        $\mathcal{Y}$ & Output space \\
        \hline
        $h$ & A classifier \\
        $\mathbf{D}_{\mathrm{train}}$& Training dataset $\{(\vec{x}_n,y_n)\;|\; 1 \leq n \leq N\}$ \\
        $\vec{x}_n$ & $n$-th training instance \\
        $N$ & Number of training data instances \\
        $\mathbf{D}_{\mathrm{test}}$& Test dataset $\{\vec{x}_t\;|\; 1 \leq t \leq T\}$ \\
        $\vec{x}_t$ & $t$-th test instance \\
        $T$ & Number of test instances \\
        \hline
        $\Tilde{\vec{x}_t}$& A counterfactual explanation of $\vec{x}_t$ \\
        $\boldsymbol{\phi}(\vec{x})$ & Feature importance vector $(\phi_1(\vec{x}), \cdots, \phi_Q(\vec{x}))$ \\
        $\mathcal{N}(\vec{x}_t)$ & The set of elements contained within the ball of radius $\epsilon$ centered at $\vec{x}_t$ \\
        $L(\vec{x}_t)$ &  Discrete local Lipschitz continuity for instance $\vec{x}_t$ \\
        \hline
        $K$ & Number of ensemble members\\
        $h_k$ & $k$-th ensemble member\\
        $p(y|h_k, \vec{x}_t)$ & Probabilistic prediction for $\vec{x}_t$ given by $h_k$ \\
        $m_t$ & A belief function defined over $2^\mathcal{Y}$ as the prediction for $\vec{x}_t$ \\
        $\mathrm{BetP}$ & The Dirac distribution associated with the prediction for a probabilistic form \\
        \hline
        $\mathrm{AU}(\vec{x}_t)$ & Aleatoric uncertainty in $\vec{x}_t$ \\
        $\mathrm{EU}(\vec{x}_t)$ & Epistemic uncertainty in $\vec{x}_t$ \\
        $\mathrm{TU}(\vec{x}_t)$ & Total uncertainty in $\vec{x}_t$\\
        $\mathrm{EC}(\vec{x}_t)$ & Epistemic certainty in $\vec{x}_t$\\  
        \hline
        $||\cdot||_p$ & The $L^p$-norm \\
        $||\cdot||_2$ & The $L^2$-norm (Euclidean)\\
        \hline
        XAI & Explainable Artificial Intelligence\\
        UQ & Uncertainty Quantification \\
        SHAP & Shapley Additive Explanations \\
        LIME &Local Interpretable Model-agnostic Explanations \\
        \hline
\end{tabular}
\caption{Notations and Acronyms}\label{tab:Notations and Acronyms}
\end{table}

\section{Description of the datasets used in the experiments}\label{appendix:datasets}

\begin{table}
    \centering
    \normalsize
    \begin{tabular}{@{}lccc@{}}
    \toprule
    Dataset & Instances & Classes & Features\\
    \midrule
    Breast Cancer & 596 & 2 & 30\\
    Ecoli & 336 & 8 & 7\\
    Glass & 214 & 6 & 9\\
    Heart & 303 & 2 & 7\\
    Ionosphere & 351 & 2 & 34\\
    Iris & 150 & 3 & 4\\
    Liver & 345 & 2 & 6\\
    Parkinson & 195 & 2 & 22\\
    Sonar & 208 & 2 & 60\\
    Wine & 178 & 3 & 13\\
    MNIST & 60000 & 10 & 28$\times$28\\
    \bottomrule\\
    \end{tabular}
    \caption{Datasets description, with number of observations, number of classes, and number of features.}\label{tab:Datasets}
\end{table}

\section{Counterfactual explanations similarity \emph{vs.} aleatoric uncertainty}\label{appendix:counterfactuals}

In this experiment, the robustness of counterfactual explanations is evaluated across multiple datasets. A total of 10 datasets from the UCI repository~\citep{Dua2019} are used, varying in size and complexity. For each dataset, the feature space is normalized, and the data is randomly split into training and test sets using a 70/30 ratio. Each experiment is repeated 100 times.

A $K$-Nearest Neighbors classifier is trained with $K = 7$ and default parameters from the scikit-learn library~\citep{sklearn2011}. Counterfactual explanations for each prediction are generated according to Equation~\eqref{eq:counterfactuals}, and the attainability of each counterfactual is computed using Equations~\eqref{eq:attaignability}.
This attainability is then correlated with the model’s aleatoric uncertainty, as estimated by the four state-of-the-art methods presented in Section~\ref{section:uq}. A statistical test is also performed, with the null hypothesis stating that there is no correlation between explanation dissimilarity and the model’s aleatoric uncertainty. This hypothesis is rejected for every dataset at a significance level of $\alpha = 0.05$, indicating a statistically significant correlation between aleatoric uncertainty and dissimilarity. Note that this test confirms the presence of a correlation but does not assess the accuracy or strength of the estimated correlation.

For the centroid-based approach, centroids are computed for each class in the training set using Euclidean distance (a neural network is not employed in this experiment, so the weight matrix is ignored). The predicted measure is determined by maximizing the criterion proposed by the authors~\citep{van-amersfoort20a} at Equation~\eqref{eq:eu_centroid}, which is a certainty measure rather than an uncertainty measure. The inverse of this value is taken as the epistemic uncertainty, and the total uncertainty is obtained as Shannon entropy of the relative certainty for all classes. Aleatoric uncertainty is obtained by subtracting the epistemic uncertainty to the total uncertainty. The length scale parameter of the method is set to $1$. The results are presented in Table~\ref{tab:similarity_centroids}.

The experiment is also conducted using the entropy-based ensemble method presented in~\citep{Shaker2020}.
A random forest with 100 estimators and a maximum depth of 4 is used. All other parameters are set to the default values of the scikit-learn library~\citep{sklearn2011}.
To compute aleatoric uncertainty, we used the entropy decomposition proposed at Equation~\eqref{eq:au_entropy}, where the total uncertainty is defined as the entropy of the posterior distribution.  The results are presented in Table~\ref{tab:similarity_ent}.

For the likelihood-based method, we used the plausibility formula provided by the authors in~\citep{Nguyen2022} to compute aleatoric uncertainty. The results are presented in Table~\ref{tab:similarity_density} only for 2-class datasets as this method is only compatible for binary classification tasks.

Finally, the local method based on the Evidential K-NN~\citep{denoeux1995} is also studied to estimate aleatoric uncertainty. The parameters for the method are the same as those used in the version presented by~\citep{Hoarau2024-ml}. For this experiment, we arbitrarily chose a number $K$ of neighbors equal to 7. Aleatoric uncertainty is computed according to the Discord at Equation~\eqref{eq:discord}, as proposed by the authors in~\citep{Hoarau2024-ml}. This result is highlighted in Table~\ref{tab:similarity_eknn}.

\begin{figure*}[htb]
\centering
\begin{minipage}{0.48\linewidth}
    \centering
    \captionsetup{type=tabular}
    \begin{tabular}{@{}lccc@{}}
    \toprule
    Dataset\; & Correlation & p-value &  Significance\\
    \midrule
    B. Cancer\; & -0.83 & $\simeq0$ & $\surd$\\
    Ecoli & -0.53 & $\simeq0$ & $\surd$\\
    Glass & -0.55 & $\simeq0$ & $\surd$\\
    Heart & -0.79 & $\simeq0$ & $\surd$\\
    Ionos. & -0.61 & $\simeq0$ & $\surd$\\
    Iris & -0.58 & $\simeq0$ & $\surd$\\
    Liver & -0.82 & $\simeq0$ & $\surd$\\
    Park. & -0.89 & $\simeq0$ & $\surd$\\
    Sonar & -0.82 & $\simeq0$ & $\surd$\\
    Wine & -0.45 & $1.42\times10^{-275}$ & $\surd$\\
    \bottomrule
    \end{tabular}
    \captionof{table}{Spearman's rank correlation of aleatoric uncertainty \emph{vs.} dissimilmarity of counterfactuals, for belief-based uncertainty quantification.}\label{tab:similarity_eknn}
\end{minipage}
\hspace{0.3cm}
\begin{minipage}{0.48\linewidth}
    \centering
    \captionsetup{type=tabular}
    \begin{tabular}{@{}lccc@{}}
    \toprule
    Dataset\; & Correlation & p-value & Significance\\
    \midrule
    B. Cancer\; & -0.52 & $\simeq 0$& $\surd$\\
    Ecoli & -0.21 & $2.14\times10^{-103}$& $\surd$\\
    Glass & -0.27 & $5.13\times10^{-117}$ & $\surd$\\
    Heart & -0.33 & $1.45\times10^{-232}$& $\surd$\\
    Ionos. & -0.04 & $5.60\times10^{-7}$&$\surd$\\
    Iris & -0.73 & $\simeq 0$&$\surd$\\
    Liver & -0.38 & $\simeq 0$&$\surd$\\
    Park. & -0.66 & $\simeq 0$&$\surd$\\
    Sonar & -0.32 & $4.56\times10^{-158}$ & $\surd$\\
    Wine & -0.49 & $\simeq 0$& $\surd$\\
    \bottomrule
    \end{tabular}
    \captionof{table}{Spearman's rank correlation of aleatoric uncertainty \emph{vs.} dissimilmarity of counterfactuals, for centroid-based uncertainty quantification.}\label{tab:similarity_centroids}
\end{minipage}
\\
\begin{minipage}{0.48\linewidth}
    \centering
    \captionsetup{type=tabular}
    \begin{tabular}{@{}lccc@{}}
    \toprule
    Dataset\; & Correlation & p-value & Significance\\
    \midrule
    B. Cancer\; & -0.29 & $\simeq0$ & $\surd$\\
    Ecoli & -0.13 & $3.69\times10^{-42}$ & $\surd$\\
    Glass & 0.15 & $4.52\times10^{-37}$ & $\surd$\\
    Heart & -0.13 & $4.32\times10^{-38}$ & $\surd$\\
    Ionos. & 0.30 & $1.69\times10^{-224}$ & $\surd$\\
    Iris & -0.50 & $4.80\times10^{-286}$ & $\surd$\\
    Liver & -0.11 & $3.43\times10^{-33}$ & $\surd$\\
    Park. & -0.69 & $\simeq0$ & $\surd$\\
    Sonar & -0.07 & $9.47\times10^{-10}$ & $\surd$\\
    Wine & -0.29 & $2.96\times10^{-112}$ & $\surd$\\
    \bottomrule
    \end{tabular}
    \captionof{table}{Spearman's rank correlation of aleatoric uncertainty \emph{vs.} dissimilmarity of counterfactuals, for ensemble-based uncertainty quantification.}\label{tab:similarity_ent}
\end{minipage}
\hspace{0.3cm}
\begin{minipage}{0.48\linewidth}
    \centering
    \captionsetup{type=tabular}
    \begin{tabular}{@{}lccc@{}}
    \toprule
    Dataset\; & Correlation & p-value & Significance\\
    \midrule
    B. Cancer\; & -0.40 & $\simeq0 $& $\surd$\\
    Heart & -0.47 & $\simeq0 $& $\surd$\\
    Ionos. & 0.16 &$2.50\times10^{-69}$&  $\surd$\\
    Liver & -0.47 & $\simeq0 $&$\surd$\\
    Park. & -0.76 &$\simeq0 $& $\surd$\\
    Sonar & -0.35 & $3.61\times10^{-183}$ &$\surd$\\
    \bottomrule
    \end{tabular}
    \captionof{table}{Spearman's rank correlation of aleatoric uncertainty \emph{vs.} dissimilmarity of counterfactuals, for likelihood-based uncertainty quantification.}\label{tab:similarity_density}
\end{minipage}
\end{figure*}

\section{Feature-attribution robustness \emph{vs.} aleatoric uncertainty}\label{appendix:feature-att}

In this experiment, we evaluate the robustness of feature-attribution explanations across ten datasets from the UCI repository~\citep{Dua2019}. For each dataset, features are normalized and data is randomly split into training and test sets using a 70/30 ratio. Each experiment is repeated 5 times (due to SHAP expensive complexity). A statistical test is also performed on the 70\% of instances with the lowest epistemic uncertainty, with the null hypothesis stating that there is no correlation between explanation robustness and the model’s aleatoric uncertainty. This hypothesis is rejected for almost every dataset at a significance level of $\alpha = 0.05$, indicating a statistically significant correlation between aleatoric uncertainty and un-robustness. Note that this test confirms the presence of a correlation but does not assess the accuracy or strength of the estimated correlation.

A $K$-Nearest Neighbors classifier with $K = 7$ is trained using default parameters from the scikit-learn library. SHAP explanations are generated for each prediction based on Equation~\eqref{eq:shap}, and their robustness is computed using Equations~\eqref{eq:robustness_level}. This robustness is then correlated with the model's aleatoric uncertainty, according to the four state-of-the-art methods introduced in Section~\ref{section:uq}.

For the centroid-based method, class centroids are computed in the training set using Euclidean distance. The predicted certainty measure is obtained by maximizing the criterion proposed in~\citep{van-amersfoort20a} at Equation~\eqref{eq:eu_centroid}. Aleatoric uncertainty is quantified by the Shannon entropy of the relative certainty across all classes. The method's length-scale parameter is fixed to 1. Results are shown in Table~\ref{tab:robustness_centroids}.

We also apply the entropy-based ensemble method from~\citep{Shaker2020}, using a random forest with 100 estimators and maximum depth of 4; other parameters follow scikit-learn defaults~\citep{sklearn2011}. To compute aleatoric uncertainty, we used the entropy decomposition proposed at Equation~\eqref{eq:au_entropy}. Results are presented in Table~\ref{tab:robustness_entropy}.

For the likelihood-based approach, aleatoric uncertainty is computed based on the plausibility function introduced by~\citep{Nguyen2022}. This method is only applicable to binary classification tasks, results are reported in Table~\ref{tab:robustness_density} only for two-class datasets.

Finally, the local Evidential K-NN method~\citep{denoeux1995} is also evaluated. Parameters for the method follow the setup described in~\citep{Hoarau2024-ml}, with $K = 7$ neighbors. Aleatoric uncertainty is assessed using both Discord~\eqref{eq:discord} and Non-specificity~\eqref{eq:nonspe} measures, as proposed by the authors. These results are presented in Table~\ref{tab:robustness_eknn}.

\begin{figure*}[htb]
\centering
\begin{minipage}{0.48\linewidth}
    \centering
    \captionsetup{type=tabular}
        \begin{tabular}{@{}lccc@{}}
        \toprule
        Dataset\;\; & Correlation & \;\;\;p-value\;\;\; & Significance\\
        \midrule
        B. Cancer & 0.43 & $8.23\times10^{-29}$ & $\surd$\\
        Ecoli & 0.57 & $6.73\times10^{-32}$ & $\surd$\\
        Glass & 0.53 & $4.53\times10^{-18}$ & $\surd$\\
        Heart & 0.48 & $2.08\times10^{-20}$ & $\surd$\\
        Ionos. & 0.18 & $3.91\times10^{-4}$ & $\surd$\\
        Iris & 0.68 & $5.50\times10^{-23}$ & $\surd$\\
        Liver & 0.50 & $1.51\times10^{-24}$ & $\surd$\\
        Park. & 0.46 & $3.83\times10^{-12}$ & $\surd$\\
        Sonar & -0.05 & $4.06\times10^{-1}$ & $\times$\\
        Wine & 0.52 & $1.29\times10^{-14}$ & $\surd$\\
        \bottomrule
    \end{tabular}
    \captionof{table}{Spearman's rank correlation of aleatoric uncertainty \emph{vs.} robustness of feature-importance explanations, for belief-based uncertainty quantification.}\label{tab:robustness_eknn}
\end{minipage}
\hspace{0.3 cm}
\begin{minipage}{0.48\linewidth}
    \centering
    \captionsetup{type=tabular}
    \begin{tabular}{@{}lccc@{}}
        \toprule
        Dataset\; & Correlation & p-value & Significance \\
        \midrule 
        B. Cancer & 0.37 & $3.45\times10^{-21}$ & $\surd$\\        
        Ecoli & 0.12 & $1.77\times10^{-2}$ & $\surd$\\
        Glass & 0.33 & $3.51\times10^{-7}$ & $\surd$\\
        Heart & 0.43 & $5.20\times10^{-16}$ & $\surd$\\
        Ionos. & 0.11 & $3.66\times10^{-2}$& $\surd$\\
        Iris & 0.26 & $9.54\times10^{-4}$& $\surd$\\
        Liver & 0.14 & $4.43\times10^{-3}$& $\surd$\\   
        Park. & 0.06 & $3.35\times10^{-1}$& $\times$\\
        Sonar & 0.19 & $3.38\times10^{-3}$& $\surd$\\
        Wine & 0.38 & $5.63\times10^{-8}$& $\surd$\\
        \bottomrule
    \end{tabular}
    \captionof{table}{Spearman's rank correlation of aleatoric uncertainty \emph{vs.} robustness of feature-importance explanations, for centroid-based uncertainty quantification.}\label{tab:robustness_centroids}
\end{minipage}
\\
\begin{minipage}{0.48\linewidth}
    \centering
    \captionsetup{type=tabular}
    \begin{tabular}{@{}lccc@{}}
        \toprule
        Dataset\; & Correlation & p-value & Significance \\
        \midrule
        B. Cancer & 0.07 & $5.68\times10^{-2}$ & $\times$ \\  
        Ecoli & 0.39 & $3.31\times10^{-20}$ & $\surd$ \\
        Glass & 0.11 & $7.70\times10^{-2}$ & $\times$\\
        Heart & 0.44 & $2.19\times10^{-16}$ & $\surd$\\
        Ionos. & 0.30 & $2.06\times10^{-9}$ & $\surd$\\
        Iris & 0.17 & $3.41\times10^{-2}$ & $\surd$\\
        Liver & 0.32 & $3.12\times10^{-10}$ & $\surd$\\   
        Park. & 0.32 & $1.64\times10^{-6}$ & $\surd$\\   
        Sonar & 0.17 & $1.06\times10^{-2}$ & $\surd$\\
        Wine & 0.39 & $2.33\times10^{-8}$ & $\surd$\\
        \bottomrule
    \end{tabular}
    \captionof{table}{Spearman's rank correlation of aleatoric uncertainty \emph{vs.} robustness of feature-importance explanations, for ensemble-based uncertainty quantification.}\label{tab:robustness_entropy}
\end{minipage}
\hspace{0.3cm}
\begin{minipage}{0.48\linewidth}
    \centering
    \captionsetup{type=tabular}
    \begin{tabular}{@{}lccc@{}}
        \toprule
        Dataset\; & Correlation & p-value &Significance\\
        \midrule
        B. Cancer & 0.03 & $3.71\times10^{-1}$ & $\times$\\
        Heart & 0.46 & $2.32\times10^{-18}$ &$\surd$\\
        Ionos. & 0.43 & $2.37\times10^{-18}$ &$\surd$\\
        Liver & 0.56 & $1.44\times10^{-31}$ &$\surd$\\   
        Park. & 0.35 &$2.10\times10^{-7}$ & $\surd$ \\
        Sonar & 0.01 & $9.86\times10^{-1}$ &$\times$ \\
        \bottomrule
    \end{tabular}
    \captionof{table}{Spearman's rank correlation of aleatoric uncertainty \emph{vs.} robustness of feature-importance explanations, for likelihood-based uncertainty quantification.}\label{tab:robustness_density}
\end{minipage}
\end{figure*}

\section{Rejection of explanations based on epistemic uncertainty}\label{appendix:rejection}

For this experiment, we illustrate the proposition of rejecting explanations based on epistemic uncertainty in both machine learning and deep learning contexts.

For the case of feature-importance explanations, we used the Glass dataset from the UCI repository~\citep{Dua2019}. After normalization, the dataset was split into training and test sets using a $(70\%, 30\%)$ ratio. A $K$-Nearest Neighbors classifier was then trained with $K = 7$ and default parameters from the scikit-learn library~\citep{sklearn2011}.
Epistemic uncertainties were computed according to Equation~\eqref{eq:nonspe} for each prediction on the test set. Only the most epistemically uncertain prediction is illustrated. For this test instance, the corresponding SHAP explanation was constructed using Equation~\eqref{eq:shap}. The remainder of this analysis, as described in the main document, shows that this explanation involves features for which the model cannot make a robust inference.

For the case of counterfactual explanations, we used MNIST~\citep{Lecun1998}, a 10-class dataset consisting of 28x28 grayscale images of handwritten digits, divided into 60,000 training images and 10,000 test images.
On this dataset, we trained LeNet~\citep{Lecun1998} with a batch size of 20. We used only 2,000 training instances in order to avoid having an overly performant model, which would not serve our purpose. After training, the test set was projected onto the extracted feature space (prior to the linear layers). Epistemic uncertainties were computed using Equation~\eqref{eq:nonspe} for each test instance. Again, only the most epistemically uncertain prediction is shown. For this test instance, the corresponding counterfactual explanation was constructed using Equation~\eqref{eq:counterfactuals}. As with the previous case, the remainder of the analysis, presented in the main document, demonstrates that such explanations do not offer additional transparency to the user for any given prediction.


\end{document}